\pdfoutput=1

\documentclass[11pt]{article}

\usepackage{acl}
\usepackage{graphicx, subfig}
\usepackage{xcolor}

\usepackage{times}
\usepackage{latexsym}
\usepackage{amsmath}
\usepackage{mathtools}
\usepackage{amsthm}
\usepackage{subcaption}
\usepackage{multirow}
\usepackage{makecell}
\usepackage[justification=justified]{caption}
\newtheorem{example}{Example}

\usepackage[T1]{fontenc}

\usepackage[utf8]{inputenc}

\usepackage{microtype}

\usepackage{inconsolata}

\usepackage{tabularray}
\UseTblrLibrary{booktabs}

\title{Timeline-based Sentence Decomposition with In-Context Learning for Temporal Fact Extraction}

\author{
    Jianhao Chen\textsuperscript{1}, Haoyuan Ouyang\textsuperscript{1}, Junyang Ren\textsuperscript{1}, Wentao Ding\textsuperscript{2}, Wei Hu\textsuperscript{1}, Yuzhong Qu\textsuperscript{1} \\
    \textsuperscript{1}\textit{State Key Laboratory for Novel Software Technology, Nanjing University, China}\\
    \textsuperscript{2}\textit{State Key Laboratory of General Artificial Intelligence, BIGAI. Beijing, China}\\
    \texttt{\{jh\_chen, hyouyang, jyren\}@smail.nju.edu.cn}\\
    \texttt{dingwentao@bigai.ai, \{whu,yzqu\}@nju.edu.cn}
}

\date{}

\begin{document}
\maketitle
\begin{abstract}
Facts extraction is pivotal for constructing knowledge graphs. Recently, the increasing demand for temporal facts in downstream tasks has led to the emergence of the task of temporal fact extraction. In this paper, we specifically address the extraction of temporal facts from natural language text. Previous studies fail to handle the challenge of establishing time-to-fact correspondences in complex sentences. To overcome this hurdle, we propose a timeline-based sentence decomposition strategy using large language models (LLMs) with in-context learning, ensuring a fine-grained understanding of the timeline associated with various facts. In addition, we evaluate the performance of LLMs for direct temporal fact extraction and get unsatisfactory results. To this end, we introduce TSDRE\footnote{The released source code and documentation are available at \url{https://github.com/JianhaoChen-nju/TSDRE}}, a method that incorporates the decomposition capabilities of LLMs into the traditional fine-tuning of smaller pre-trained language models (PLMs). To support the evaluation, we construct ComplexTRED, a complex temporal fact extraction dataset. Our experiments show that TSDRE achieves state-of-the-art results on both HyperRED-Temporal and ComplexTRED datasets.

\end{abstract}

\section{Introduction}
Acquiring knowledge has long been a fundamental challenge in the field of artificial intelligence. Typically, the acquired knowledge is stored in knowledge graphs (KGs) as triples, comprising a head entity, a relation, and a tail entity. Recently, there has been a significant surge in the need~\cite{MGTQA,LGTQA,DING,TKGR,TIRGN,Recom,xiong2024large} for acquiring temporal facts. For instance, 
users may seek information from the KGs, such as ``When did Michael Jordan win the NBA Finals MVP?''. However, traditional triples like (NBA Finals MVP, winner, Michael Jordan) cannot meet such needs due to the lack of time dimension. This can be solved by adding time information like (Point In Time, 1996) to the triple. A temporal fact can be formalized as a quintuple ($e_{head}$, $r$, $e_{tail}$, $q$, $t$), where the time qualifier $q$ and time value $t$ indicate the time dimension information of the triple.

\begin{figure}
    \centering
    \includegraphics[width=\columnwidth]{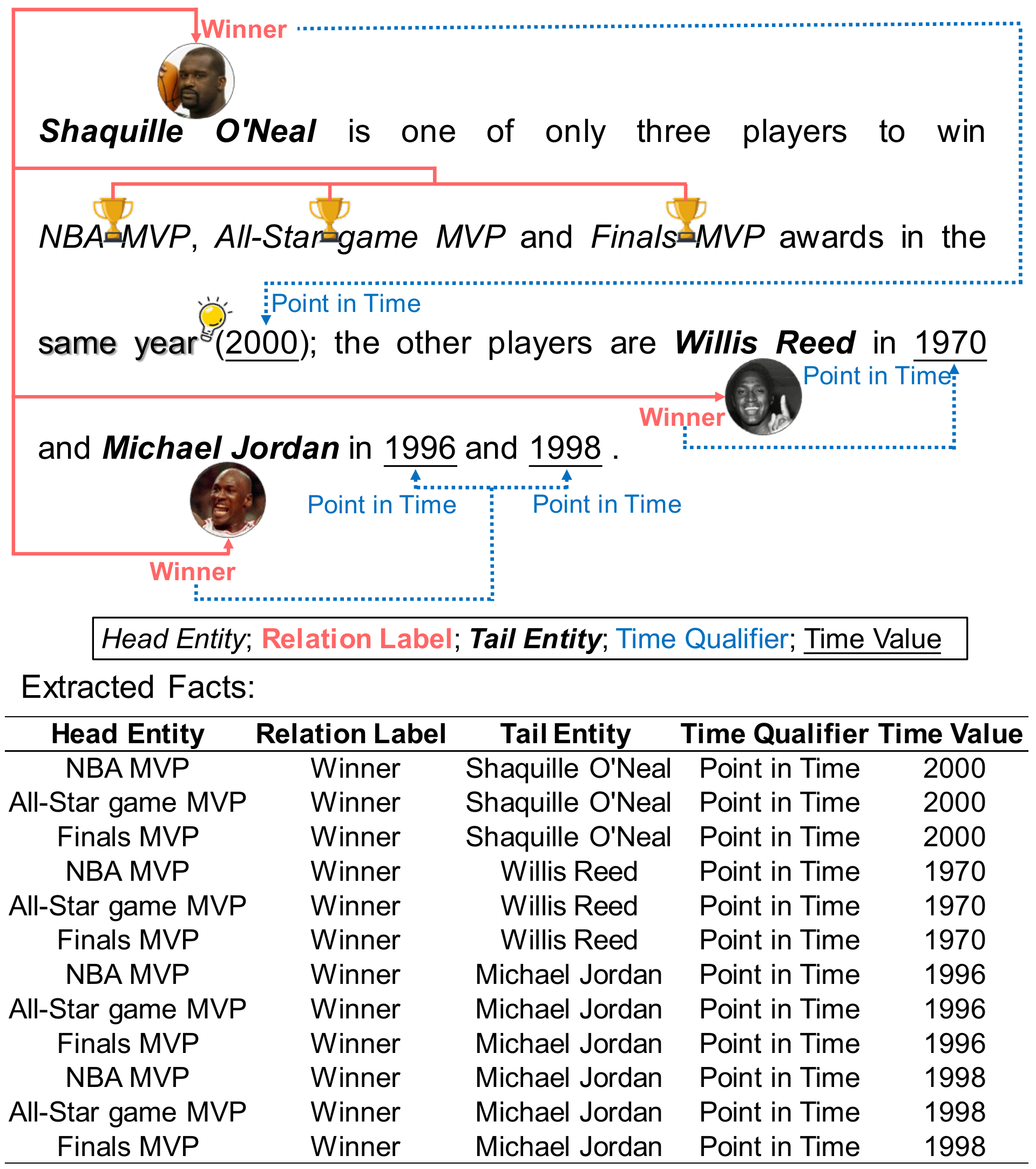}
    \caption{A difficult temporal fact extraction example, which contains 12 temporal facts in one sentence.}
    \label{fig:complexInstance}
\end{figure}

Although many works~\cite{pravda,WWWJ,CubeRE} have explored the task of temporal fact extraction, the challenge of establishing correspondence between time and triples remains unresolved. Figure~\ref{fig:complexInstance} shows a difficult example of temporal fact extraction. It can be seen that the difficulty of this example mainly comes from the use of ``the other" to imply that the other two players also won three awards in one year. Besides, natural language expressions are typically succinct, with "Michael Jordan in 1996 and 1998" alone expressing six temporal facts. Existing methods cannot handle this situation without explicitly stating what event occurred at a certain point in time, leading to suboptimal performance in addressing complex narratives with interwoven timelines.

To address this gap, we propose a timeline-based sentence decomposition strategy. Our strategy involves breaking down sentences according to their timelines to capture the temporal dimensions of facts, ensuring a fine-grained understanding of the timeline associated with various facts. In the past, sentence decomposition often required a large amount of training corpus. But now, the in-context learning capability of large language models (LLMs) empowers us to perform timeline-based sentence decomposition without training corpus. 

To the best of our knowledge, we are the first to investigate the application of LLMs for temporal fact extraction tasks. We conduct an evaluation to assess the performance of employing LLMs directly for temporal fact extraction. However, it is unsatisfactory that the performance of LLMs does not surpass the traditional approach of fine-tuning smaller PLMs. To enhance the performance of temporal fact extraction methods, we come up with the idea of combining the surprising sentence decomposition capability of LLMs with the traditional way of fine-tuning small pre-trained language models. Experiments show that our method based on the combining idea achieves SOTA results.

In addition to research on extraction methods, there are currently few benchmarks~\cite{CubeRE,pravdaDataset} for temporal fact extraction, and they do not pay much attention to sentences involving complex time related narratives (we call them complex sentences for short). To this end, we construct a temporal fact extraction dataset composed of complex sentences for evaluation. 

In summary, our main contributions in this paper are outlined as follows:
\begin{enumerate}
\item We summarize the unique challenge of the temporal fact extraction task and propose a timeline-based sentence decomposition method on natural language using in-context learning enhanced by human feedback.
\item We propose timeline-based sentence decomposition (TSD) for the temporal fact extraction task. Evaluation results demonstrate that TSD indeed helps models understand the correspondence between events and time.
\item We conduct evaluations of the methods that utilize LLMs in the task of temporal extraction. Moreover, we build a novel dataset ComplexTRED, which consists of 19,148 complex sentences with multiple time expressions or temporal facts.
\end{enumerate}


\section{Related Work}
\label{sec:relatedwork}
\subsection{Temporal Fact Extraction}
Temporal fact extraction methods are mainly divided into two categories: pattern-based methods and deep learning-based methods.

\paragraph{Pattern-based methods}
Existing work attempts to treat relationships and qualifiers as a whole to mine corresponding patterns in text, or use sequence annotation and classification methods. T-Yago~\cite{TYAGO} and a similar study~\cite{TempWeb} extract temporal instances from semi-structured data like Wikipedia's Infoboxes, Categories, and Lists, limiting coverage to freely available text. Pravda~\cite{pravda} uses textual patterns to represent candidate facts and labels them through a graph-based label propagation algorithm. \citet{WWWJ} applies the idea of distant supervision, leveraging existing temporal facts to learn corresponding patterns from web text and subsequently applying them to the extraction process.

\paragraph{Deep learning-based methods}
CubeRE~\cite{CubeRE} first employs a sequence labeling approach to identify entities and time, followed by classification of the relations and qualifiers between them.

Previous Studies have not fully addressed the challenges posed by complex sentences in temporal fact extraction. Moreover, we are the first to explore the application of LLMs for the task of temporal fact extraction.

\subsection{Sentence Decomposition}
Sentence decomposition is a common method used for tasks in the NLP field. From a technical perspective, sentence decomposition can be roughly divided into two categories: supervised learning methods~\cite{QDT} and rule-based methods~\cite{EDG}. Supervised learning requires sufficient training corpus, while rule-based methods are labor-intensive and often suffer from poor coverage issues. In the era of large models, we explore using in-context learning for sentence decomposition, which does not require task-specific training data and has sufficient coverage.

\subsection{In-Context Learning}
Large language models, exemplified by the GPT series~\cite{Brown2020,OpenAI} and the LlaMA~\cite{Llama,llama2} family, are known to have impressive in-context learning abilities. These LLMs have been proven to be able to solve a completely new problem with a small number of examples without the need for task-specific training leading to a surge in exploration within the domain of in-context learning.

Generally, research on in-context learning can be categorized into two main areas. The first focuses on the strategy for selecting examples~\cite{select} to enhance performance, while the second delves into the interpretability aspects~\cite{interpret} of in-context learning. Our study focuses on example selection. We iteratively constructed demonstrative examples for task scenarios without training corpus. Moreover, we incorporate human feedback to guide the model and prevent common errors.

\section{Preliminary}
\label{sec:preli}
\subsection{Problem Formulation}
\label{sec:problemform}
Given an input sentence of $n$ words $s$ = \{$x_1$, $x_2$, ..., $x_n$\}, the objective of the temporal fact extraction task is to extract all existing temporal facts in $s$. Formally, a temporal fact is represented as ($e_{head}$, $r$, $e_{tail}$, $q$, $t$). An entity $e$ is a consecutive span of words where $e$ = \{$x_i$, $x_{i+1}$, ..., $x_j$\}, $i, j\in$ \{$1$, ..., $n$\}. $r$ represents the relation between head entity $e_{head}$ and tail entity $e_{tail}$. $r \in R$, where $R$ is the predefined set of relation labels. The qualifier $q$ and the time value $t$ indicates the time dimension of the relation triplet ($e_{head}$, $r$, $e_{tail}$). $q \in Q$, where $Q$ is the predefined set of qualifier labels. 


\begin{figure*}[!h]
    \centering
    \includegraphics[width=\textwidth]{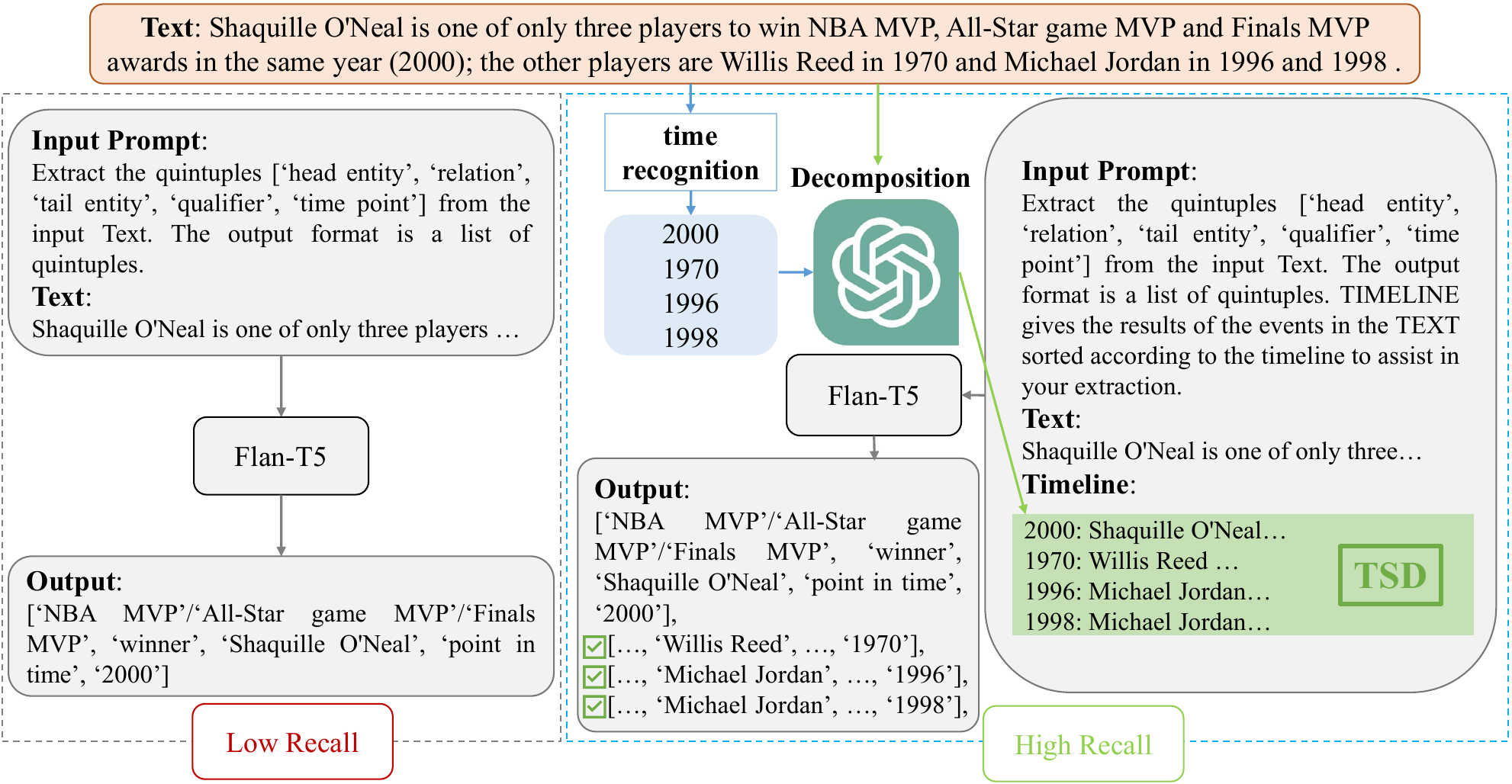}
    \caption{Framework comparison between Flan-T5 and TSDRE. Leveraging Timeline-based sentence decomposition for training can significantly improve the recall of the Flan-T5.}
    \label{fig:framework}
\end{figure*}

\subsection{Complex Sentence in Temporal Fact Extraction}
\label{sec:complexSentence}
In this paper, we delineate the distinctions between simple and complex sentences in the context of temporal fact extraction. Prior research has shown a lack of emphasis on complex sentences. However, this paper shifts its focus to the temporal fact extraction of complex sentences. 

A simple sentence comprises only one time element and one temporal fact, presenting relatively lower difficulty in relation extraction and time selection. Example \ref{eg:simplesentence} provides a concrete instance of a simple sentence in the context of temporal fact extraction.
\begin{example}\label{eg:simplesentence}
Peter Whittle was elected a fellow of the Royal Society in 1981 .

Extracted Temporal Facts:

(Peter Whittle, member of, Royal Society, start time, 1981)
\end{example}

On the other hand, a complex sentence involves two or more time elements or two or more facts, often introducing challenges in time selection or relation extraction. Understanding the correspondence between time and facts becomes more challenging in the presence of multiple time references compared to a situation with a single time reference. Figure~\ref{fig:complexInstance} has shown a complex sentence with more than two time elements.

Example~\ref{eg:complexsentence} provides another concrete instance of a complex sentence in the context of temporal fact extraction.
\begin{example}\label{eg:complexsentence}
20 November 1883, Jules Ferry succeeds Challemel-Lacour as Minister of Foreign Affairs.

Extracted Temporal Facts:

(Jules Ferry, position held, Minister of Foreign Affairs, start time, 20 November 1883)

(Challemel-Lacour, position held, Minister of Foreign Affairs, end time, 20 November 1883)

(Jules Ferry, replaces, Challemel-Lacour, point in time, 20 November 1883)
\end{example}
Although there is only one point in time in Example~\ref{eg:complexsentence}, this sentence contains three temporal facts. This is mainly because the word "succeeds" expresses the relationship of succession at this moment. Implicit expressions such as these imply the connection of events in the time dimension, which brings challenges to the complete extraction of temporal facts.

\section{Method}
In this section, we begin by presenting two direct approaches to leverage LLMs for temporal fact extraction. Following that, we introduce our extraction methods, encompassing a timeline-based sentence decomposition strategy and the fine-tuning of generative models using decomposition results for training.
\label{sec:method}
\subsection{In-Context Learning with ChatGPT3.5 and Fine-tuning Open-Source LLM}
We try to employ LLMs directly for temporal fact extraction through in-context learning and fine-tuning.

Specifically, we apply in-context learning to ChatGPT3.5. To construct the prompt, we first give a task description: \texttt{Extract all the quintuples [subject, relation, object, qualifier, time point] from the input text}, followed by the specified relationship list and qualifier list. We then select 48 examples at random from the train set to ensure that all relations and qualifiers have at least one-shot examples. The complete prompt is more than 2,000 tokens and will be shown in Appendix~\ref{sec:appendixE}.

We also try to LoRA fine-tune Llama2 (7B). We still give Llama2 the task description as the instruction to make Llama2 better understand the task.

\subsection{Timeline-based Sentence Decomposition}
Timeline-based sentence decomposition can be used to understand and process texts containing temporal information. It helps organize and present the timeline in sentences, making it easier for models to understand and follow the development of events.

An example of decomposition is as follows:

\noindent\textbf{\texttt{Text:}}
\texttt{Shaquille O'Neal is one of only three players to win NBA MVP, All-Star game MVP, and Finals MVP awards in the same year (2000); the other players are Willis Reed in 1970 and Michael Jordan in 1996 and 1998.}

\noindent\textbf{\texttt{Time:}}
\texttt{[`2000', `1970', `1996', `1998']}

\noindent\textbf{\texttt{Decomposition:}}
\texttt{2000: Shaquille O'Neal is one of only three players to win NBA MVP, All-Star game MVP, and Finals MVP awards in the same year.
1970: Willis Reed won NBA MVP, All-Star game MVP, and Finals MVP awards in the same year. 
1996: Michael Jordan won NBA MVP, All-Star game MVP, and Finals MVP awards in the same year.
1998: Michael Jordan won NBA MVP, All-Star game MVP, and Finals MVP awards in the same year.}
\begin{figure}[tb]
    \centering
    \includegraphics[width=0.48\textwidth]{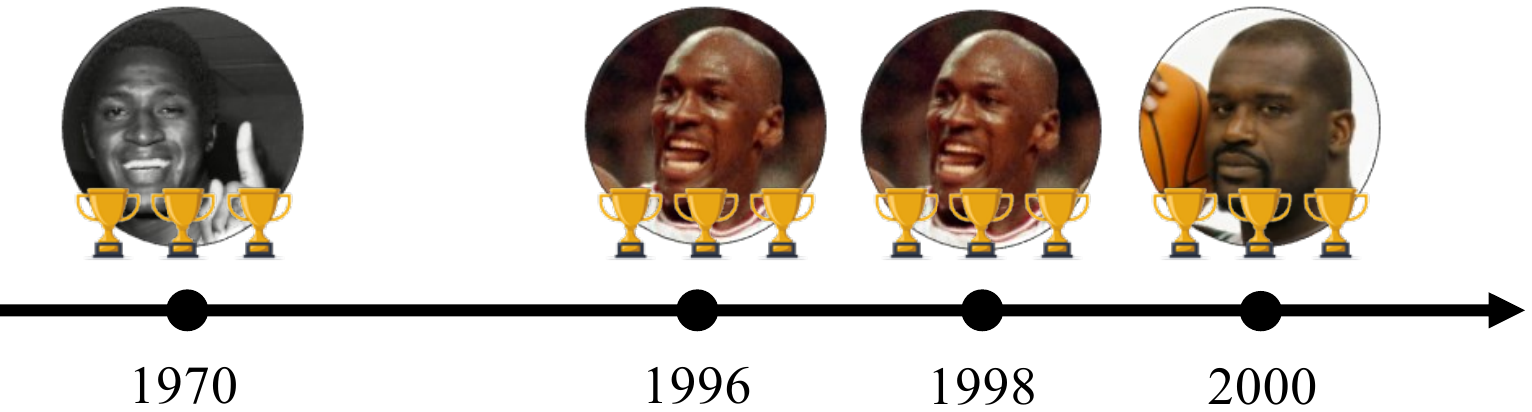}
    \caption{Player awards are presented in a timeline.}
    \label{fig:decomposition_timeline}
\end{figure}

Due to the absence of annotated data, we consider leveraging the in-context learning capability of LLMs to accomplish the decomposition task. Before starting in-context learning, we use SUTime~\cite{sutime} to identify time expressions in sentences, as specialized tools excel in time recognition compared to large LLMs. Without even a decomposition example, we iteratively construct the prompt for in-context learning. First, we only give ChatGPT3.5 the task description and one test sentence:

\noindent\textbf{\texttt{Instruction:}}
\texttt{First, I will give you a Text, and secondly, I will give you the time contained in the Text. You need to sort out the events that occurred at each time I provided based on the Text. The Decomposition format is required to contain a time point and corresponding events in each sentence. Each sentence contains the complete elements of the event, such as subject, predicate, and object.}

We undertake this exploration to understand the output format preferences of ChatGPT3.5. This knowledge will guide us in uncovering the potential of the model to achieve optimal performance by aligning with these preferences.
Subsequently, we make slight modifications to the output to generate a demonstrative example that is both accurate and in line with the model's preferred output format. After several iterations like this, we get the demonstrative examples needed for in-context learning.

\paragraph{Human Feedback Enhanced In-Context Learning}
We find that even though we have shown ChatGPT3.5 high-quality demonstrative examples, it still makes mistakes when outputting. We add negative examples in the prompt to help the model avoid common mistakes. We carefully select the examples that we believe are representative of the mistakes made by ChatGPT3.5. Subsequently,  we add human feedback to point out the mistakes and provide corrected answers. The complete prompt will be shown in Appendix~\ref{sec:appendixE}.

\subsection{Fine-tuning Models with Timeline-based Sentence Decomposition}

We combine the natural language understanding and reasoning capabilities emerging from very large LMs with smaller LMs fine-tuned for specific tasks. Specifically, we fine-tune generative language models (Llama2 and Flan-T5) with timeline-based sentence decomposition. We splice the text in the dataset with the decomposition results generated by ChatGPT3.5 as a new input for training. The following is an example of the input we provide for models:

\noindent\textbf{\texttt{Input:}}
\texttt{Text: Lamberto Visconti di Eldizio ( died 1225 ) was the Judge of Gallura from 1206 , when he married the heiress Elena , to his own death . Decomposition: 1225: Lamberto Visconti di Eldizio passed away and ended serving as the Judge of Gallura. 1206: Lamberto Visconti di Eldizio became the Judge of Gallura and married the heiress Elena.}

Figure~\ref{fig:framework} illustrates the distinction between training directly and training with the utilization of temporal-based sentence decomposition (TSD) information in the framework. We name the method of fine-tuning Flan-T5 using TSD as TSDRE.

\section{ComplexTRED: A Complex Temporal Fact Extraction Dataset}
\label{sec:dataset}
There are very few temporal fact extraction datasets available to date. Pravda \cite{pravdaDataset} lacks annotation labels and therefore cannot be used for supervised learning. The Wiki-People Dataset~\cite{WWWJ} is not open source. HyperRED \cite{CubeRE} is a hyper-relational fact extraction dataset, 48\% of which are temporal facts. However, many samples in HyperRED are overly simplistic and insufficient to depict the potential difficulties in complex practical scenarios. Specifically, the majority of sentences in HyperRED consist solely of a single temporal expression, and their respective extraction results comprise a singular temporal fact.

We need a complex temporal fact extraction dataset to evaluate our method and train existing models on their ability to extract temporal facts from complex temporal sentences. Below we will introduce data collection and dataset statistics.

\subsection{Data Collection}
It is very challenging to construct a large-scale, diverse complex temporal fact extraction dataset. The two major difficulties are the number of complex temporal sentences and high-quality annotated labels. 
In order to collect enough data, we use two methods to obtain data: using distant supervision to obtain the alignment of web text and temporal fact in KG, and manually correcting the samples in HyperRED. We will introduce our means of controlling the quality of the dataset in the introduction of each method.

\paragraph{Distant Supervision}
We collect the introduction sections of 343,603 DBpedia articles, the full text of 401,796 Wikipedia articles, and 3,002,373 Wikidata temporal facts for alignment. Specifically, we use DBpedia Spotlight~\cite{spotlight} for entity linking and SUTime~\cite{sutime} to extract time entities from DBpedia and Wikidata articles. In distant supervision, if a sentence contains the head entity, tail entity and temporal entity of a temporal fact, then we align the fact to the sentence.

Distant supervision encounters two significant challenges: noise and incomplete facts (The sentence may express facts that are not in the knowledge graph). However, by aligning the two entities plus time, the noise problem is greatly mitigated. To tackle the incomplete facts problem, we leverage ChatGPT to complete the facts contained in the sentence. Subsequently, we organize 50 computer science undergraduate students to manually label the correctness of the added facts by ChatGPT3.5. Finally, we obtain about 17,000 complex sentences through distant supervision.

\paragraph{Manual Correction from HyperRED}
Samples in HyperRED are alignment with Wikipedia text and Wikidata data, which are from the same source as our samples. Therefore, we correct the labels of some complex sentences from HyperRED and absorb them into our dataset.  We first select about 7,000 sentences in HyperRED that contain more than or equal to two time expressions as candidates. Then we organize 70 computer science undergraduates to correct the labels of each sentence, and the corrected samples will be added to our dataset. Finally, we get 2,589 corrected samples.

\subsection{Dataset Statistics}
To ensure that the train set, dev set, and test set data adhere to an independent and identical distribution, we employ stratified sampling on the original data based on the relation types. Our goal is to maintain an 8:1:1 ratio for each relation type in the train, dev, and test set, respectively. The resulting splitting of the dataset is presented in Table~\ref{tab:dataset}. We manually check all samples in the dev and test set to ensure the dataset's quality.


\begin{table}[tb]
    \centering
    {\small
    \begin{tblr}{
        width=\columnwidth,
        colspec = { X[l, 1.5] X[l] X[r] X[r] X[r] },
        row{1} = {font=\bfseries}
    }
    \toprule
    \SetCell[c=2]{c} Datasets & & \#Sent. & \#Facts & \#Rel.\\
    \midrule
    HyperR. & Train & 17,004 & 23,507 & 45 \\
    & Dev & 432 & 582 & 37 \\
    & Test & 1,712 & 2,391 & 41 \\
    \midrule
    ComplexT. & Train & 16,573 & 33,632 & 40 \\
    & Dev & 1,679 & 4,025 & 40 \\
    & Test & 1,584 & 3,964 & 39 \\
    \bottomrule
    \end{tblr}}
    \caption{\label{tab:dataset} Statistics about the number of sentences, temporal facts, and relations of the two datasets.}
\end{table}

The analysis of the sample is shown in the Table~\ref{tab:datasetAnalysis}. ComplexTRED significantly outperforms HyperRED-Temporal in terms of sentence length, temporal expressions, and temporal facts contained per sentence, which reflects the difficulty of the dataset to a certain extent.
\begin{table}[tb]
\centering
{\small
    \begin{tblr}{
        width=\columnwidth,
        colspec = { X[l, 1.5] X[r] X[r] },
        row{1} = {font=\bfseries},
        cell{1}{1}={c}
    }
    \toprule
    Metrics & HyperR. & ComplexT. \\
    \midrule
    Sentence length & 31.47 & 41.27  \\
    Time expressions & 1.95 & 3.46 \\
    Temporal facts& 1.38 & 2.10  \\
    \bottomrule
    \end{tblr}}
\caption{\label{tab:datasetAnalysis}
Sample-level comparison between HyperRED-Temporal and ComplexTRED. Each metric is calculated on an average per sample basis.
}

\end{table}


\section{Experiments}
\label{sec:experiment}

\begin{table*}[htb]
    \centering
    {\small
    \begin{tblr}{
        width=\textwidth,
        colspec = { X[8, l] X[r] X[r] X[r] X[r] X[r] X[r] },
        row{1} = {font=\bfseries},
        cell{1}{1}={c}
    }
    \toprule
    \SetCell[r=2]{c} Methods  & \SetCell[c=3]{c} HyperRED-Temporal & & & \SetCell[c=3]{c} ComplexTRED & & \\
    \cmidrule{2-4} \cmidrule{5-7}
    & Pr & Re & F$_1$  & Pr & Re & F$_1$\\
    \toprule
    ChatGPT3.5$^\dagger$ & 11.26 & 20.08 & 14.43 & 19.01 & 23.57  & 21.05 \\
    Llama2 w LoRA$^\dagger$~\cite{llama2} & 56.95 & 31.87 & 40.87 & 35.72 & 17.74 & 23.71 \\
    Llama2 w LoRA$^\dagger$ + TSD$^\dagger$ (OURS) & 56.60 & 48.06 & 51.98  & 40.64 & 27.18 & 32.58 \\
    CubeRE~\cite{CubeRE} & 55.31 & 49.64 & 52.33 & 41.69 & 27.92 & 33.44 \\
    Flan-T5~\cite{flan-t5} & \underline{66.64} & \underline{61.06} & \underline{63.73} & 47.64 & \underline{34.86} & \underline{40.26} \\
    Flan-T5 + Explanation$^\dagger$~\cite{Revisiting} & 64.99 & 58.85 & 61.76 & \underline{47.69} & 34.07 & 39.75 \\
    TSDRE$^\dagger$ w Flan-T5 (OURS) & \textbf{68.61} & \textbf{64.91} & \textbf{66.71} & \textbf{48.99} & \textbf{37.61} & \textbf{42.55} \\
    \bottomrule
    \end{tblr}}
    \caption{Main results on HyperRED-Temporal and ComplexTRED. $\dagger$ denotes that the corresponding foundation model is a LLM.}
    \label{tab:mainresult}
\end{table*}


\subsection{Experimental Setup}
\label{sec:setting}
\paragraph{Datasets}
We assess temporal fact extraction using HyperRED~\cite{CubeRE}, a publicly available benchmark, and ComplexTRED, a dataset specifically crafted by us for complex temporal fact extraction. We select samples in HyperRED whose labels are temporal facts, delete the relations that have little correlation with time, and finally form HyperRED-Temporal as the evaluation dataset. The statistics of the two datasets are shown in Table~\ref{tab:dataset}.

\paragraph{Compared Methods}
We compare our approach with the SOTA extraction method reported on CubeRE~\cite{CubeRE} and several baseline methods.
We evaluated the performance of directly employing LLMs for temporal fact extraction, including LoRA fine-tuning the 7B version of Llama2~\cite{llama2} and naive in-context learning with ChatGPT3.5. We also tried to leverage our Timeline-based Sentence Decomposition (TSD) to enhance LoRA fine-tuned Llama2.

In addition to directly employing LLMs, we also evaluated the performance of Flan-T5-Large~\cite{flan-t5}, a relatively smaller pre-trained language model (PLM). We transplant the method of \citet{Revisiting} on relation extraction (triple extraction) to temporal fact extraction, which trains Flan-T5 with LLM-generated explanations. We use ChatGPT3.5 to generate the explanations required for training instead of GPT-3 in the~\citet{Revisiting}, because ChatGPT3.5 has not yet come out when~\citet{Revisiting} was published and we believe ChatGPT3.5 performs better than GPT-3.

\paragraph{Evaluation metrics}
We report the precision (denoted as $P$), recall (denoted as $R$), and $F_1$ score of the evaluation results. When calculating metrics, we employ an exact match (at string level) approach without following \citet{Revisiting}'s method of utilizing manual assessment for the outputs of large-scale models.


\subsection{Main results}
\label{sec:mainresults}
We first report the results of methods directly based on LLMs, and subsequently, we report the results of approaches that involve combining LLMs and smaller PLMs.

\paragraph{LLMs results.} Based on Table~\ref{tab:mainresult}, it is evident that both Fine-tuned open-source LLMs and in-context learning LLMs do not perform satisfactorily on both datasets. Among them, in-context ChatGPT3.5 consistently gets the lowest $F_1$ score. It is reasonable to achieve such performance without training. HyperRED-Temporal includes 45 relation types and 4 qualifier types, while ComplexTRED includes 40 relation types and 4 qualifier types. Despite providing detailed task settings for ChatGPT3.5, it is expected to be exceedingly challenging to provide accurate answers across all 180 (45*4)/160 (40*4) categories based solely on a few-shot examples. Fine-tuned Llama2 (7B) performs better than in-context ChatGPT3.5, but its $F_1$ score still falls considerably short of CubeRE. This could be attributed to the insufficient training samples to effectively support such a large language model, leading to suboptimal fitting. Additionally, we conducted experiments using our decomposition method to enhance Llama2 training. This resulted in Llama2 achieving an 11-point $F_1$ score improvement on HyperRED-Temporal, and a 9-point $F_1$ score improvement on ComplexTRED. This result indicates that integrating our decomposition method into training enables the model to better learn the relationships between sentence features and temporal facts. Another possible reason we speculate is that high-quality decomposition results are suitable as training data and make the training of Llama2 more sufficient.

\paragraph{Results of combining LLMs and smaller PLMs.} Flan-T5 (Large) achieved surprising results on both datasets. We believe that compared to LLMs, smaller PLMs are easier to fine-tune and fit when the training data is insufficient. However, enhancing Flan-T5 with ChatGPT3.5-generated explanation resulted in a tiny drop in $F_1$ on both datasets. This method may be disadvantageous under exact match, which we have mentioned in the evaluation metrics in Section~\ref{sec:setting}. Another reason we believe is that the temporal fact extraction task is inherently more challenging to interpret compared to the fact (triple) extraction task. Finally, our method TSDRE, which enhances Flan-T5 with timeline-based sentence decomposition (TSD), achieves state-of-the-art results on both datasets. We achieve remarkable results by combining the strengths of both large and small LMs: Large LMs excel at effectively organizing timelines in natural language, while small LMs prove more adept at precise fine-tuning for specific tasks.

\subsection{Decomposition Quality}
We randomly select 100 sentences and invite three experts to evaluate our decomposition results of these sentences. We still use precision and recall as evaluation metrics. The goal of our decomposition is to divide different events in the text into different points in time. In this scenario, True Positive refers to the count of events correctly classified at their respective time points in the prediction results. False Positive indicates the count of events erroneously classified at the wrong time point. False Negative represents the count of events that were not predicted.
It must be pointed out that the relatively ambiguous aspect is that opinions differ among individuals regarding what should be considered an event in natural language text.
\begin{table}[htb]
\centering
{\small
\begin{tabular}{lrr}
\toprule
& \textbf{Precision} & \textbf{Recall} \\
\midrule
Prompt & 93.80 & 93.53  \\
Prompt + feedback & \textbf{94.65} & \textbf{95.57} \\
\bottomrule
\end{tabular}}
\caption{\label{tab:decompositionQuality}
Human evaluation of Timeline-based sentence decomposition results.
}
\end{table}

Table~\ref{tab:decompositionQuality} shows the scores of manual evaluation. Overall, our decomposition results surpass 90 in both precision and recall metrics, indicating that we have acquired high-quality auxiliary information for training. This establishes a solid foundation for enhancing model performance. Moreover, prompt enhanced by human feedback has further improved precision and recall. This shows that human feedback on demonstrative examples can enable LLMs to understand the task requirement better and avoid some common errors.

\subsection{Error Analysis}
We select 50 sentences from each of the two datasets whose $F_1$ scores are less than 1 to analyze TSDRE's performance. The results are illustrated in Table~\ref{tab:error}.
\begin{table}[htb]
\centering
{\small
\begin{tabular}{lrr}
\toprule
\textbf{Main Errors} & \textbf{HyperR.} & \textbf{ComplexT.} \\
\midrule
NER & 14\% & 24\% \\
~~- totally wrong &8\% &12\% \\
~~- overlapped &6\% &12\% \\
Relation Extraction &20\%  &22\%   \\
Qualifier Classification &0\% &0\% \\
Time Selection &2\%  &6\%   \\
False Negative &30\% &22\% \\
Missing Facts &34\%  &26\% \\
\bottomrule 
\end{tabular}}
\caption{\label{tab:error}
The statistics of main errors of sampled sentences.
}
\end{table}
We classify error types according to the elements of the quintuple as Named Entity Recognition (NER) error, Relation Extraction error, Time Selection error, Qualifier Classification error, False Negative and Missing Facts. It is worth noting that the error rates of Time selection and Qualifier Classification are very low, showing that TSDRE perform well when faced with time-to-fact correspondences. In addition, under relaxed standards, the prediction results of entities overlapping with the answer entities and the prediction results of false negatives can both be counted as correct. This means that the actual performance of the model is much better than the scores under the exact match measurement. Finally, completely wrong NER, wrong Relation Extraction, and Missing Facts are still the legacy problems of fact (triple) extraction.

\subsection{Case Study}
As is shown in Figure~\ref{fig:framework}, Flan-T5 fails to capture information regarding the awards of the other two players, aside from O'Neal. However, after we incorporate decomposition into training, Flan-T5 successfully
outputs all temporal facts. Smaller language models do have limited learning capabilities for implicit expressions, so introducing LLMs with powerful understanding capabilities for natural language can effectively make up for this shortcoming.

\section{Conclusion}
\label{sec:conclusion}
In this paper, we explore the application of large language models (LLMs) in the extraction of temporal facts. Our attempts indicate that directly employing LLMs for temporal fact extraction falls short of achieving satisfactory results. 
To tackle this issue, we introduce a timeline-based sentence decomposition (TSD) method. Building upon this, we propose TSDRE, which employs a relatively smaller PLM as its foundation, combined with LLM-driven TSD to achieve the extraction.
Experiments demonstrate that TSDRE achieves SOTA results on two datasets and incorporating TSD into the training process can enhance the performance of LLMs on temporal fact extraction tasks.
In the future, an interesting topic would be to explore the extraction of temporal facts from text that necessitate inferring the occurrence time based on existing temporal references, such as ``three days later'', which has not yet received widespread attention.

\section*{Acknowledgments}
This work was supported by the National Natural Science Foundation of China (No. 62272219) and the Collaborative Innovation Center of Novel Software Technology \& Industrialization.

\section*{Limitations}
Our contribution does have important limitations. First, our decomposition results rely on ChatGPT for completion, and decomposition without training the open-source LLMs cannot achieve the desired results.

Second, we only tested the effect of ChatGPT on the GPT3.5-turbo model, but not on the latest GPT4 or GPT4-turbo, due to the significantly higher cost involved.

Third, for document-level temporal fact extraction, when combined with time-based sentence decomposition results, the input may exceed the maximum length allowed by the generative model.

Finally, our dataset construction inevitably introduces noise problems due to the use of distant supervision. Additionally, due to limited resources, we only checked the validation set and test set of ComplexTRED, which may result in some noise issues in the training set.

\section*{Ethics Statement}
First and foremost, our proposed TSDRE method strives to enhance the overall performance of RE models in extracting temporal facts. However, given the inherent black-box nature of the generative model, it is inevitable that the extracted facts may possess certain quality issues. Hence, when employing our method to extract temporal facts and utilize them for downstream tasks, users must exercise caution in discerning the authenticity of these facts in order to mitigate potential real-world consequences arising from erroneous information.

Secondly, for dataset construction, we have gathered text from Wikipedia and DBpedia, as well as facts from Wikidata. These are publicly available datasets commonly utilized for dataset construction. Wikidata facts are under the Creative Commons CC0 License\footnote{\url{https://www.wikidata.org/wiki/Wikidata:Licensing##Uses}}, while the texts obtained from both Wikipedia and DBpedia are licensed under the Creative Commons Attribution-ShareAlike 3.0 Unported License\footnote{\url{https://www.dbpedia-spotlight.org/licenses}}\footnote{\url{https://en.wikipedia.org/wiki/Wikipedia:Copyrights}}. Thus, we are able to freely utilize this data to construct our dataset, and our dataset will be released under the same license. Furthermore, we have organized some human annotators throughout the dataset construction process, and each annotator has been duly compensated based on their respective working hours.

\bibliography{custom}

\appendix

\section{Environments and Parameters}
\label{sec:appendixA}
TSDRE’s results were achieved using a Python implementation
running on a workstation with an Intel(R) Xeon(R) Gold 5222 CPU @ 3.80GHz, 376GB RAM and 3 NVIDIA RTX3090 graphics cards. Llama2's results were achieved on a workstation with an Intel(R) Xeon(R) Gold 6248 CPU @ 2.50GHz, 472GB RAM, and 8 NVIDIA Tesla V100 graphics cards. The hyperparameter settings of CubeRE, Flan-T5, Llama2 and BART are shown in Table~\ref{model repro}.  When applying LoRA to fine-tune Llama2, we used a rank of 8 and an alpha value of 32. Besides, We set the temperature value of ChatGPT3.5 to 0 to facilitate reproduction.

\begin{table*}[htb]
\centering
{\small
\begin{tabular}{lccccc}
\toprule
\textbf{Models} & \textbf{Data} & \textbf{Batch Sizes} & \textbf{Warm-up} 
& \textbf{Learning Rates} & \textbf{Max Epochs}\\
\midrule
\multirow{2}{*}{CubeRE} & HyperRED & 32 & 0.2 & 5e-5 & 30 \\
    & ComplexTRED & 32 & 0.2 & 5e-5 & 30 \\
\midrule
\multirow{2}{*}{Flan-T5} & HyperRED & 2 & 0.12 & 2e-5 & 4 \\
    & ComplexTRED & 2 & 0.12 & 2e-5 & 4 \\
\midrule
\multirow{2}{*}{Llama2} & HyperRED & 4 & default & 1e-4 & 3 \\
    & ComplexTRED & 4 & default & 1e-4 & 3 \\
\midrule
\multirow{2}{*}{BART} & HyperRED & 2 & 0.12 & 2e-5 & 4 \\
    & ComplexTRED & 2 & 0.12 & 2e-5 & 4 \\
\bottomrule
\end{tabular}}
\caption{Hyperparameters for CubeRE, Flan-T5, Llama2 and BART.}
\label{model repro}
\end{table*}

\section{TSDRE with BART}
\label{sec:appendixB}
In this paper, we propose a pipeline architecture TSDRE which is backbone-free. We have reported in Section 6.2 that our model TSDRE w Flan-T5 achieves the SOTA results. We test the performance of replacing Flan-T5 Large (770M) with a slightly weaker generative model BART Large (340M). The results are shown in table~\ref{tab:bart}. From the experimental results, it can be seen that the performance of BART as base model is indeed lower than that of Flan-T5 (mentioned in Table~\ref{tab:mainresult}). However, TSD still enhances the performance of BART.

\begin{table*}[htb]
    \centering
    {\small
    \begin{tblr}{
        width=\textwidth,
        colspec = { X[8, l] X[r] X[r] X[r] X[r] X[r] X[r] },
        row{1} = {font=\bfseries},
        cell{1}{1}={c}
    }
    \toprule
    \SetCell[r=2]{c} Methods  & \SetCell[c=3]{c} HyperRED-Temporal & & & \SetCell[c=3]{c} ComplexTRED & & \\
    \cmidrule{2-4} \cmidrule{5-7}
    & Pr & Re & F$_1$  & Pr & Re & F$_1$\\
    \toprule
    BART Large (340M)~\cite{bart} & 68.85 & 48.89 & 57.18 & 43.68 & 17.52 & 25.00 \\
    TSDRE$^\dagger$ w BART Large (340M) & \textbf{71.51} & \textbf{50.27} & \textbf{59.04} & \textbf{45.15} & \textbf{18.45} & \textbf{26.20} \\
    \bottomrule
    \end{tblr}}
    \caption{Performance of BART as base model on HyperRED-Temporal and ComplexTRED. $\dagger$ denotes that the corresponding foundation model is a LLM.}
    \label{tab:bart}
\end{table*}

\section{Error Analysis}
\label{sec:appendixC}
Here we present specific examples of three major types of errors.

\subsection{NER Error}
\noindent\textbf{\texttt{Input:}}
\texttt{Fifteen locomotives of British Rail Class 83 were built between 1960 and 1962 by English Electric at Vulcan Foundry , as part of British Rail 's policy to develop a standard electric locomotive .}

\noindent\textbf{\texttt{Prediction:}}
\texttt{[['British Rail Class 83', 'manufacturer', 'Vulcan Foundry', 'end time', '1962'], ['British Rail Class 83', 'manufacturer', 'Vulcan Foundry', 'start time', '1960']]}

\noindent\textbf{\texttt{Error Analysis:}}
\texttt{The tail entity should be 'English Electric'.}

\subsection{Relation Extraction Error}
\noindent\textbf{\texttt{Input:}}
\texttt{He was appointed manager of the Highland Railway ' s Lochgorm Works in 1903 , and promoted to Assistant to the Chief Mechanical Engineer , Peter Drummond in 1903 .}

\noindent\textbf{\texttt{Prediction:}}
\texttt{[['Peter Drummond', 'employer', 'Highland Railway', 'start time', '1903'], ['Peter Drummond', 'employer', 'Highland Railway', 'end time', '1903']]}

\noindent\textbf{\texttt{Error Analysis:}}
\texttt{Relation should be 'director/ manager'.}

\subsection{Missing Facts}
\noindent\textbf{\texttt{Input:}}
\texttt{Sir Hubert Edward Henry Jerningham , ( 18 October 1842 - 3 April 1914 ) was a British Liberal Party politician and Governor of Mauritius 1892 - 1897 , then Governor of Trinidad and Tobago between 1897 and 1900 .}

\noindent\textbf{\texttt{Prediction:}}
\texttt{[['Hubert Edward Henry Jerningham', 'position held', 'Governor of Trinidad and Tobago','start time', '1897'], ['Hubert Edward Henry Jerningham', 'position held', 'Governor of Mauritius', 'end time', '1897'], ['Hubert Edward Henry Jerningham', 'position held', 'Governor of Mauritius','start time', '1892']]}

\noindent\textbf{\texttt{Error Analysis:}}
\texttt{['Hubert Edward Henry Jerningham', 'position held', 'Governor of Trinidad and Tobago','end time', '1900'] missed.}

\section{Difference from the temporal relation extraction task}
\label{sec:appendixD}
Indeed, it is easy to confuse temporal relation extraction with temporal fact extraction based on their names, but they are actually very different tasks. Specifically, temporal relation extraction aims to identifying the temporal relation (e.g., BEFORE, AFTER, OVERLAPS) between events and times, while temporal fact extractions (please refer to Section~\ref{sec:problemform}) aims to to extract facts with temporal attributes (e.g., start\_time, end\_time). Therefore, datasets used for temporal relation extraction tasks such as TimeBank~\cite{TimeBank} and TempEval~\cite{tempeval} are not suitable for evaluating our task.

\section{Prompt}
\label{sec:appendixE}
\subsection{In-Context ChatGPT3.5 Prompt}
\noindent\textbf{\texttt{Task:}}
\texttt{Extract all the quintuples [subject, relation, object, qualifier, time point] from the input text.}

\noindent\textbf{\texttt{Task requirements:}}
\texttt{Extract all the quintuples [subject, relation, object, qualifier, time point] from the input text.\\
Here are some concrete examples, the output format is a list of quintuples:}

\noindent\textbf{\texttt{Input:}}
\texttt{He received the 1921 Nobel Prize in Physics for his \" services to theoretical physics \" , in particular his discovery of the law of the photoelectric effect , a pivotal step in the evolution of quantum theory .}

\noindent\textbf{\texttt{Output:}}
\texttt{[['Nobel Prize', 'winner', 'He', 'point in time', '1921'], ['He', 'award received', 'Nobel Prize', 'point in time', '1921']]}

\noindent\textbf{\texttt{Input:}}
\texttt{It won the 1991 Nebula Award for Best Novelette and was nominated for the 1991 Hugo Award for Best Novelette .}

\noindent\textbf{\texttt{Output:}}
\texttt{[['It', 'award received', 'Nebula Award', 'point in time', '1991'], ['It', 'nominated for', 'Best Novelette', 'point in time', '1991']]}

\noindent\textbf{\texttt{Input:}}
\texttt{They are currently the only club in Ulster to have won an All - Ireland Senior Club Hurling Championship , which they first won in 1983 .}

\noindent\textbf{\texttt{Output:}}
\texttt{[['All - Ireland Senior Club Hurling Championship', 'winner', 'They', 'point in time', '1983']]}

\noindent\textbf{\texttt{Input:}}
\texttt{Alexander Mackenzie , PC ( January 28 , 1822 \u2013 April 17 , 1892 ) , was a building contractor and newspaper editor , and was the second Prime Minister of Canada , from November 7 , 1873 to October 8 , 1878 .}

\noindent\textbf{\texttt{Output:}}
\texttt{[['Canada', 'head of government', 'Alexander Mackenzie', 'end time', 'October 8 , 1878'], ['Canada', 'head of government', 'Alexander Mackenzie', 'start time', 'November 7 , 1873'], ['Alexander Mackenzie', 'position held', 'Prime Minister', 'end time', 'October 8 , 1878'], ['Alexander Mackenzie', 'position held', 'Prime Minister', 'start time', 'November 7 , 1873']]}

\noindent\textbf{\texttt{Input:}}
\texttt{There has been a resident Treasury or Downing Street cat employed as a mouser and pet since the reign of Henry VIII , when Cardinal Wolsey placed his cat by his side while acting in his judicial capacity as Lord Chancellor , an office he assumed in 1515 .}

\noindent\textbf{\texttt{Output:}}
\texttt{[['Cardinal Wolsey', 'position held', 'Lord Chancellor', 'start time', '1515']]}

\noindent\textbf{\texttt{Input:}}
\texttt{He had 24 caps for Japan , from 1974 to 1984 , scoring 3 tries , 5 conversions , 14 penalties and 3 drop goals , in an aggregate of 73 points .}

\noindent\textbf{\texttt{Output:}}
\texttt{[['He', 'member of sports team', 'Japan', 'start time', '1974'], ['He', 'member of sports team', 'Japan', 'end time', '1984']]}

\noindent\textbf{\texttt{Input:}}
\texttt{He later played professional football in the American Football League , appearing in 42 games as a tackle and defensive end for the New York Titans ( later renamed the Jets ) from 1960 to 1962 .}

\noindent\textbf{\texttt{Output:}}
\texttt{[['He', 'member of sports team', 'New York Titans', 'start time', '1960'], ['He', 'member of sports team', 'New York Titans', 'end time', '1962']]}

\noindent\textbf{\texttt{Input:}}
\texttt{" Suedehead " is the debut solo single from Morrissey , released in February 1988 .}

\noindent\textbf{\texttt{Output:}}
\texttt{[['Suedehead', 'performer', 'Morrissey', 'publication date', 'February 1988']]}

\noindent\textbf{\texttt{Input:}}
\texttt{It closed on 1 December 2003 when operation of the line was suspended between Kabe Station and Sandanky\u014d Station .}

\noindent\textbf{\texttt{Output:}}
\texttt{[['It', 'adjacent station', 'Kabe', 'end time', '1 December 2003'], ['It', 'adjacent station', 'Sandanky\u014d', 'end time', '1 December 2003']]}

\noindent\textbf{\texttt{Input:}}
\texttt{The 2008 presidential campaign of Barack Obama , then junior United States Senator from Illinois , was announced at an event on February 10 , 2007 in Springfield , Illinois .}

\noindent\textbf{\texttt{Output:}}
\texttt{[['Barack Obama', 'candidacy in election', '2008 presidential campaign', 'start time', 'February 10 , 2007']]}

\noindent\textbf{\texttt{Input:}}
\texttt{Among his victories were in reconquering Ji ' an in Jiangxi Province in 1856 , as well as leading the assault on the Taiping capital at Nanjing in 1864 .}

\noindent\textbf{\texttt{Output:}}
\texttt{[['Nanjing', 'capital of', 'Taiping', 'end time', '1864']]}

\noindent\textbf{\texttt{Input:}}
\texttt{He was also one of the original correspondents on Comedy Central ' s The Daily Show from 1996 to 1998 .}

\noindent\textbf{\texttt{Output:}}
\texttt{[['The Daily Show', 'cast member', 'He', 'end time', '1998'], ['The Daily Show', 'cast member', 'He', 'start time', '1996']]}

\noindent\textbf{\texttt{Input:}}
\texttt{He was International President of WWF from 1996 to 1999 succeeding Prince Philip , the Duke of Edinburgh .}

\noindent\textbf{\texttt{Output:}}
\texttt{[['WWF', 'chairperson', 'He', 'end time', '1999'], ['WWF', 'chairperson', 'He', 'start time', '1996']]}

\noindent\textbf{\texttt{Input:}}
\texttt{Timothy Fok Tsun - ting ( born 14 February 1946 in Hong Kong ) , GBS , JP , the eldest son of Henry Fok , is a Member of the Legislative Council of Hong Kong , representing the Sports , Performing Arts , Culture and Publication functional constituency .}

\noindent\textbf{\texttt{Output:}}
\texttt{[['Henry Fok', 'child', 'Timothy Fok Tsun - ting', 'start time', '14 February 1946']]}

\noindent\textbf{\texttt{Input:}}
\texttt{The 2013 Philadelphia Eagles season was the franchise ' s 81st season in the National Football League , and the first under head coach Chip Kelly .}

\noindent\textbf{\texttt{Output:}}
\texttt{[['Chip Kelly', 'coach of sports team', 'Philadelphia Eagles', 'start time', '2013']]}

\noindent\textbf{\texttt{Input:}}
\texttt{Shek Kip Mei Station served as a terminus in the very early phase of the Kwun Tong Line ( Shek Kip Mei to Kwun Tong , 1 October 1979 to 31 December 1979 ) .}

\noindent\textbf{\texttt{Output:}}
\texttt{[['Shek Kip Mei', 'connecting line', 'Kwun Tong Line', 'start time', '1 October 1979']]}

\noindent\textbf{\texttt{Input:}}
\texttt{K\u00f6nigsberg was transferred to Soviet control in 1945 after World War II .}

\noindent\textbf{\texttt{Output:}}
\texttt{[['K\u00f6nigsberg', 'country', 'Soviet', 'start time', '1945']]}

\noindent\textbf{\texttt{Input:}}
\texttt{She was the Director of the Walter and Eliza Hall Institute of Medical Research ( WEHI ) , from 1996 until 30 June 2009 and remains a faculty member , having rejoined the institute ' s Molecular Genetics of Cancer Division .}

\noindent\textbf{\texttt{Output:}}
\texttt{[['WEHI', 'director / manager', 'She', 'end time', '30 June 2009'], ['WEHI', 'director / manager', 'She', 'start time', '1996']]}

\noindent\textbf{\texttt{Input:}}
\texttt{He previously served as Commander , United States Transportation Command from September 2005 to August 2008 .}

\noindent\textbf{\texttt{Output:}}
\texttt{[['United States Transportation Command', 'director / manager', 'He', 'end time', 'August 2008'], ['United States Transportation Command', 'director / manager', 'He', 'start time', 'September 2005']]}

\noindent\textbf{\texttt{Input:}}
\texttt{He graduated from Pennsylvania State University in State College , PA in 1969 , and earned a J . D .}

\noindent\textbf{\texttt{Output:}}
\texttt{[['He', 'educated at', 'Pennsylvania State University', 'end time', '1969']]}

\noindent\textbf{\texttt{Input:}}
\texttt{She married in 1936 , and took up her first post in Liverpool University , where she studied for the rest of her working life .}

\noindent\textbf{\texttt{Output:}}
\texttt{[['She', 'employer', 'Liverpool University', 'start time', '1936']]}

\noindent\textbf{\texttt{Input:}}
\texttt{Air Union was merged with four other French airlines to become Air France on 7 October 1933 .}

\noindent\textbf{\texttt{Output:}}
\texttt{[['Air Union', 'followed by', 'Air France', 'point in time', '7 October 1933']]}

\noindent\textbf{\texttt{Input:}}
\texttt{The Duchy of Magdeburg ( German : Herzogtum Magdeburg ) was a province of Brandenburg - Prussia from 1680 to 1701 and a province of the German Kingdom of Prussia from 1701 to 1807 .}

\noindent\textbf{\texttt{Output:}}
\texttt{[['Brandenburg - Prussia', 'followed by', 'Prussia', 'point in time', '1701'], ['Magdeburg', 'located in the administrative territorial entity', 'Prussia', 'start time', '1701'], ['Magdeburg', 'located in the administrative territorial entity', 'Brandenburg - Prussia', 'start time', '1680']]}

\noindent\textbf{\texttt{Input:}}
\texttt{In 1998 , the studio moved from Studio City , California to Burbank in celebration of a new facility , and was renamed Nickelodeon Animation Studio .}

\noindent\textbf{\texttt{Output:}}
\texttt{[['Nickelodeon Animation Studio', 'headquarters location', 'Studio City , California', 'end time', '1998'], ['Nickelodeon Animation Studio', 'headquarters location', 'Burbank', 'start time', '1998']]}

\noindent\textbf{\texttt{Input:}}
\texttt{Kimera Walusimbi was Kabaka of the Kingdom of Buganda between 1374 and 1404 .}

\noindent\textbf{\texttt{Output:}}
\texttt{[['Kimera', 'noble title', 'Kabaka', 'start time', '1374'], ['Kimera', 'noble title', 'Kabaka', 'end time', '1404']]}

\noindent\textbf{\texttt{Input:}}
\texttt{For the start of the 1982 season , the Minnesota Vikings moved from Metropolitan Stadium to the Hubert H . Humphrey Metrodome .}

\noindent\textbf{\texttt{Output:}}
\texttt{[['Metrodome', 'occupant', 'Minnesota Vikings', 'start time', '1982'], ['Minnesota Vikings', 'home venue', 'Metrodome', 'start time', '1982']]}

\noindent\textbf{\texttt{Input:}}
\texttt{It was established as the official legislature of Kampuchea on January 5 , 1976 , consisting of 250 members .}

\noindent\textbf{\texttt{Output:}}
\texttt{[['Kampuchea', 'legislative body', 'It', 'start time', 'January 5 , 1976']]}

\noindent\textbf{\texttt{Input:}}
\texttt{They were designed by R . J . Billinton and built at Brighton works from 1895 to 1897 .}

\noindent\textbf{\texttt{Output:}}
\texttt{[['They', 'manufacturer', 'Brighton works', 'start time', '1895']]}

\noindent\textbf{\texttt{Input:}}
\texttt{Dee Palmer ( formerly David Palmer ; born 2 July 1937 ) is an English composer , arranger , and keyboardist best known for having been a member of the progressive rock group Jethro Tull from 1977 to 1980 .}

\noindent\textbf{\texttt{Output:}}
\texttt{[['David Palmer', 'member of', 'Jethro Tull', 'end time', '1980'], ['David Palmer', 'member of', 'Jethro Tull', 'start time', '1977']]}

\noindent\textbf{\texttt{Input:}}
\texttt{In 1931 , he joined Joseph Lyons and several other members in leaving the Labor Party and joining with the Nationalists to create the United Australia Party .}

\noindent\textbf{\texttt{Output:}}
\texttt{[['Joseph Lyons', 'member of political party', 'United Australia Party', 'start time', '1931'], ['Joseph Lyons', 'member of political party', 'Labor Party', 'end time', '1931']]}

\noindent\textbf{\texttt{Input:}}
\texttt{He was in the United States Army during World War II , from 1943 to 1946 .}

\noindent\textbf{\texttt{Output:}}
\texttt{[['He', 'military branch', 'United States Army', 'end time', '1946'], ['He', 'military branch', 'United States Army', 'start time', '1943']]}

\noindent\textbf{\texttt{Input:}}
\texttt{She was named after the title character of the 1866 opera Mignon , written by her godfather , French composer Ambroise Thomas .}

\noindent\textbf{\texttt{Output:}}
\texttt{[['Ambroise Thomas', 'notable work', 'Mignon', 'publication date', '1866']]}

\noindent\textbf{\texttt{Input:}}
\texttt{It was best known as the home of the Detroit Red Wings hockey team of the National Hockey League from its opening until 1979 .}

\noindent\textbf{\texttt{Output:}}
\texttt{[['Detroit Red Wings', 'home venue', 'It', 'end time', '1979'], ['It', 'occupant', 'Detroit Red Wings', 'end time', '1979']]}

\noindent\textbf{\texttt{Input:}}
\texttt{He became a solicitor in 1900 and a barrister in 1913 , being a member of both King ' s Inns , Dublin , and Gray \u2019 s Inn , London .}

\noindent\textbf{\texttt{Output:}}
\texttt{[['He', 'occupation', 'solicitor', 'start time', '1900'], ['He', 'occupation', 'barrister', 'start time', '1913']]}

\noindent\textbf{\texttt{Input:}}
\texttt{The Portuguese Air Force ( PoAF ) operated 50 LTV A - 7 Corsair II aircraft in the anti - ship , air interdiction and air defense roles between 1981 and 1999 .}

\noindent\textbf{\texttt{Output:}}
\texttt{[['LTV A - 7 Corsair II', 'operator', 'Portuguese Air Force', 'start time', '1981'], ['LTV A - 7 Corsair II', 'operator', 'Portuguese Air Force', 'end time', '1999']]}

\noindent\textbf{\texttt{Input:}}
\texttt{Brickleberry is an American animated comedy that premiered on September 25 , 2012 on Comedy Central .}

\noindent\textbf{\texttt{Output:}}
\texttt{[['Brickleberry', 'original broadcaster', 'Comedy Central', 'start time', 'September 25 , 2012']]}

\noindent\textbf{\texttt{Input:}}
\texttt{Shellen joined Google in 2003 when the company acquired Pyra Labs , which developed the Blogger blogging platform .}

\noindent\textbf{\texttt{Output:}}
\texttt{[['Pyra Labs', 'owned by', 'Google', 'start time', '2003']]}

\noindent\textbf{\texttt{Input:}}
\texttt{Volkswagen purchased the Bugatti trademark in June 1998 and incorporated Bugatti Automobiles S . A . S .}

\noindent\textbf{\texttt{Output:}}
\texttt{[['Bugatti', 'owned by', 'Volkswagen', 'start time', 'June 1998']]}

\noindent\textbf{\texttt{Input:}}
\texttt{In 1986 , the company was acquired by Penguin Group and split into two imprints : Dutton and Dutton Children ' s Books .}

\noindent\textbf{\texttt{Output:}}
\texttt{[['Dutton', 'parent organization', 'Penguin Group', 'start time', '1986'], ['Dutton Children ' s Books', 'parent organization', 'Penguin Group', 'start time', '1986']]}

\noindent\textbf{\texttt{Input:}}
\texttt{In 1990 , following the Iraqi invasion of Kuwait , Saudi Arabia participated in the Gulf War to expel Iraqi forces from the country .}

\noindent\textbf{\texttt{Output:}}
\texttt{[['Gulf War', 'participant', 'Saudi Arabia', 'point in time', '1990']]}

\noindent\textbf{\texttt{Input:}}
\texttt{On November 18 , 1928 the first Mickey Mouse cartoon released to the public , Steamboat Willie , debuted at the Colony .}

\noindent\textbf{\texttt{Output:}}
\texttt{[['Mickey Mouse', 'present in work', 'Steamboat Willie', 'point in time', 'November 18 , 1928']]}

\noindent\textbf{\texttt{Input:}}
\texttt{The Third Republic of South Korea was replaced in 1972 by the Fourth Republic of South Korea under the Third Republic of South Korea ' s president Park Chung - hee .}

\noindent\textbf{\texttt{Output:}}
\texttt{[['Fourth Republic of South Korea', 'replaces', 'Third Republic of South Korea', 'point in time', '1972']]}

\noindent\textbf{\texttt{Input:}}
\texttt{He returned to favor in 1942 and was recalled to Moscow .}

\noindent\textbf{\texttt{Output:}}
\texttt{[['He', 'residence', 'Moscow', 'start time', '1942']]}

\noindent\textbf{\texttt{Input:}}
\texttt{On 20 July 2012 , the Constable welcomed the Olympic Torch to London at the Tower one week in advance of the London 2012 Summer Olympic Games , as part of the Olympic torch relay .}

\noindent\textbf{\texttt{Output:}}
\texttt{[['London', 'significant event', 'London 2012 Summer Olympic Games', 'point in time', '20 July 2012']]}

\noindent\textbf{\texttt{Input:}}
\texttt{and Max Verstappen , who in 2015 became the youngest driver in Formula One history at just 17 years old .}

\noindent\textbf{\texttt{Output:}}
\texttt{[['Max Verstappen', 'sport', 'Formula One', 'start time', '2015']]}

\noindent\textbf{\texttt{Input:}}
\texttt{She was married to internationally famous writer Jorge Amado from 1945 until his death in 2001 .}

\noindent\textbf{\texttt{Output:}}
\texttt{[['She', 'spouse', 'Jorge Amado', 'start time', '1945'], ['She', 'spouse', 'Jorge Amado', 'end time', '2001']]}

\noindent\textbf{\texttt{Input:}}
\texttt{It was first listed on the London Stock Exchange in 2005 is now a constituent of the FTSE 100 Index .}

\noindent\textbf{\texttt{Output:}}
\texttt{[['It', 'stock exchange', 'London Stock Exchange', 'start time', '2005']]}

\noindent\textbf{\texttt{Input:}}
\texttt{The Valencia Street Circuit ( Valencian : Circuit Urb\u00e0 de Val\u00e8ncia , Spanish : Circuito Urbano de Valencia ) was a street circuit in Valencia , Spain which hosted the Formula One European Grand Prix for five years ( 2008 \u2013 2012 ) .}

\noindent\textbf{\texttt{Output:}}
\texttt{[['Valencia Street Circuit', 'used by', 'Formula One', 'start time', '2008'], ['Valencia Street Circuit', 'used by', 'Formula One', 'end time', '2012']]}

Now please extract the quintuples from the following input:

\noindent\textbf{\texttt{Input:}}

\subsection{Decomposition Prompt with Feedback}

\texttt{First, I will give you a TEXT, and secondly, I will give you the TIME contained in the TEXT. You need to sort out the events that occurred at each time I provided based on the TEXT. The DECOMPOSITION format is required to contain a time point and corresponding events in each sentence. Each sentence contains the complete elements of the event, such as subject, predicate, and object. DECOMPOSITION ends with </s>.
There are two evaluation criteria for this task: Precision and Recall. The definitions of Precision and Recall under this task are as follows:}

\texttt{Precision: Whether the time at which each event occurred is given correctly.}

\texttt{Recall: Whether all events that occurred at each time are given completely.}

\texttt{I will show you some examples. At the same time, i will give you human feedback for decomposition results in these examples. I need you to learn from these examples and feedback how to decompose TEXT correctly and completely:}

\noindent EXAMPLE1:

\noindent\textbf{\texttt{TEXT:}}
\texttt{The Duchy of Magdeburg ( German : Herzogtum Magdeburg ) was a province of Brandenburg - Prussia from 1680 to 1701 and a province of the German Kingdom of Prussia from 1701 to 1807 .}

\noindent\textbf{\texttt{TIME:}}
\texttt{['from 1680 to 1701', 'from 1701 to 1807']}

\noindent\textbf{\texttt{DECOMPOSITION:}}
\texttt{From 1680 to 1701: The Duchy of Magdeburg was a province of Brandenburg-Prussia.
From 1701 to 1807: The Duchy of Magdeburg was a province of the German Kingdom of Prussia.
</s>}

\noindent\textbf{\texttt{feedback:}}
\texttt{Precision: very good.}
\texttt{Recall: very good.}

\noindent EXAMPLE2:

\noindent\textbf{\texttt{TEXT:}}
\texttt{From its foundation in 1921 until moving to Ashton Gate in 2014 the Memorial Ground was the home to Bristol Rugby .}

\noindent\textbf{\texttt{TIME:}}
\texttt{['1921', '2014']}

\noindent\textbf{\texttt{DECOMPOSITION:}}
\texttt{1921: The Memorial Ground was founded and became the home to Bristol Rugby.
2014: The Memorial Ground moved to Ashton Gate.
</s>}

\noindent\textbf{\texttt{feedback:}}
\texttt{Precision: Not good. It was not the Memorial Ground that was founded in 1921 but Bristol Rugby. Besides, it was not the Memorial Ground but Bristol Rugby that moved to Ashton Gate.}
\texttt{Recall: Not good. Two events are not recalled: In 1921 Bristol Rugby was founded and in 2014 Bristol Rugby moved to Ashton Gate.
A correct and complete decomposition is as follows:
1921: Bristol Rugby was founded and the Memorial Ground the home to Bristol Rugby.
2014: Bristol Rugby moved to Ashton Gate.}

\noindent EXAMPLE3:

\noindent\textbf{\texttt{TEXT:}}
\texttt{Alexander Mackenzie , PC ( January 28 , 1822 \u2013 April 17 , 1892 ) , was a building contractor and newspaper editor , and was the second Prime Minister of Canada , from November 7 , 1873 to October 8 , 1878 .}

\noindent\textbf{\texttt{TIME:}}
\texttt{['January 28, 1822', 'November 7, 1873', 'October 8, 1878', 'April 17, 1892']}

\noindent\textbf{\texttt{DECOMPOSITION:}}
\texttt{January 28, 1822: Alexander Mackenzie, PC, was born.
November 7, 1873: Alexander Mackenzie started serving as the second Prime Minister of Canada.
October 8, 1878: Alexander Mackenzie ended his term as the second Prime Minister of Canada.
April 17, 1892: Alexander Mackenzie passed away.
</s>}

\noindent\textbf{\texttt{Feedback:}}
\texttt{Precision: very good.}
\texttt{Recall: very good.}

\noindent \texttt{EXAMPLE4:}

\noindent\textbf{\texttt{TEXT:}}
\texttt{Pope Benedict XV ( Latin : Benedictus XV ; Italian : Benedetto XV ) born Giacomo Paolo Giovanni Battista della Chiesa , ( 21 November 1854 \u2013 22 January 1922 ) , was Pope from 3 September 1914 until his death in 1922 .}

\noindent\textbf{\texttt{TIME:}}
\texttt{['21 November 1854', '22 January 1922', '3 September 1914', '1922']}

\noindent\textbf{\texttt{DECOMPOSITION:}}
\texttt{21 November 1854: Pope Benedict XV, born Giacomo Paolo Giovanni Battista della Chiesa, was born.
22 January 1922: Pope Benedict XV passed away.
3 September 1914: Pope Benedict XV started serving as Pope.
1922: Pope Benedict XV ended serving as Pope.
</s>}

\noindent\textbf{\texttt{feedback:}}
\texttt{Precision: very good.}
\texttt{Recall: not good. In 22 January 1922, Pope Benedict XV passed away and was no longer Pope. In 1922: Pope Benedict XV passed away and ended serving as Pope.
A correct and complete decomposition is as follows:
21 November 1854: Pope Benedict XV, born Giacomo Paolo Giovanni Battista della Chiesa, was born.
22 January 1922: Pope Benedict XV passed away and ended serving as Pope.
3 September 1914: Pope Benedict XV started serving as Pope.
1922: Pope Benedict XV passed away and ended serving as Pope.}

\noindent EXAMPLE5:

\noindent\textbf{\texttt{TEXT:}}
\texttt{He received : Serbian NIN Prize 1978 and Andri\u0107 Award 1987 , Swedish PEN Tucholsky Award ( 1993 ) , German Herder Prize ( 1995 ) , Montenegrin Njego\u0161 Award ( 2009 ) and 13th July award ( 2004 ) , Slovenian Vilenica Award ( 2003 ) , Croatian Vladimir Nazor Award ( 2008 ) etc .}

\noindent\textbf{\texttt{TIME:}}
\texttt{['1978', '1987', '1993', '1995', '2003', '2004', '2008', '2009']}

\noindent\textbf{\texttt{DECOMPOSITION:}}
\texttt{1978: He received the Serbian NIN Prize.
1987: He received the Andrić Award.
1993: He received the Swedish PEN Tucholsky Award.
1995: He received the German Herder Prize.
2003: He received the Slovenian Vilenica Award.
2004: He received the 13th July award.
2008: He received the Croatian Vladimir Nazor Award.
2009: He received the Montenegrin Njegoš Award.
</s>}

\noindent\textbf{\texttt{feedback:}}
\texttt{Precision: very good.}
\texttt{Recall: very good.}

\noindent EXAMPLE6:

\noindent\textbf{\texttt{TEXT:}}
\texttt{It is one of the most famous old districts in Poland today and was the center of Poland ' s political life from 1038 until King Sigismund III Vasa relocated his court to Warsaw in 1596 .}

\noindent\textbf{\texttt{TIME:}}
\texttt{['1038', '1596']}

\noindent\textbf{\texttt{DECOMPOSITION:}}
\texttt{1038: It was the center of Poland's political life.
1596: King Sigismund III Vasa relocated his court to Warsaw.
</s>}

\noindent\textbf{\texttt{feedback:}}
\texttt{Precision: very good.}
\texttt{Recall: not good. In 1596 it was no longer the center of Poland ' s political life.
A correct and complete decomposition is as follows:
1038: It was the center of Poland's political life.
1596: It was no longer the center of Poland ' s political life since King Sigismund III Vasa relocated his court to Warsaw.}

\noindent EXAMPLE7:

\noindent\textbf{\texttt{TEXT:}}
\texttt{He played for Gloucestershire between 1920 and 1921 .}

\noindent\textbf{\texttt{TIME:}}
\texttt{['1920', '1921']}

\noindent\textbf{\texttt{DECOMPOSITION:}}
\texttt{1920: He played for Gloucestershire.
1921: He played for Gloucestershire.
</s>}

\noindent\textbf{\texttt{feedback:}}
\texttt{Precision: not good. This decomposition does not convey the meaning that he started playing for Gloucestershire in 1920 and ended playing for Gloucestershire in 1921.}
\texttt{Recall: very good.
A correct and complete decomposition is as follows:
1920: He started playing for Gloucestershire.
1921: He ended playing for Gloucestershire.}

\texttt{Now process the following TEXT and follow my examples above for the DECOMPOSITION format:}

\section{Relations Deleted from HyperRED}
\label{sec:appendixF}

There are 57 types of relationships with time qualifier in the HyperRED data set. We have deleted the following relations: 

\textbf{Distant supervision of these relations results in excessive noise: } 

Shares\_Border\_With, 
Country\_Of\_Citizenship, 
Instance\_Of, 
League, 
Subclass\_Of, 
Part\_Of, 
Partner\_In\_Business\_Or\_Sport.

\textbf{Easily confused relations: }

Head\_Of\_State (Head\_Of\_Government), 
Location (Located\_In\_The\_Administrative\_Territorial\_Entity), 
Part\_Of (Member\_of), 
Participating\_team (Participant), 
Voice\_Actor (Performer).

In ComplexTRED, we delete 5 relations from HyperRED-Temporal:

\textbf{Distant supervision of these relations results in excessive noise: } 

Followed\_By (Replaces), 
Award\_Received (Winner), 
Head\_Of\_Government (Position held),

\textbf{Easily confused relations: }

Connecting\_Line, 
Stock\_Exchange.

\end{document}